\documentclass[10pt,letterpaper]{article}
\usepackage{marvosym} 
\usepackage{cogsci}
\usepackage{pst-all}
\usepackage{pslatex}
\usepackage{apacite}
\usepackage{graphicx}
\usepackage{amsmath}
\usepackage{color}

\title{Abductive, Causal, and Counterfactual Conditionals\\ Under Incomplete Probabilistic Knowledge}
 
\author{{\large \bf Niki Pfeifer (niki.pfeifer@lmu.de)} \\
Munich Center for Mathematical Philosophy, LMU Munich, Germany
  \AND {\large \bf Leena Tulkki (leena.tulkki@helsinki.fi)} \\
   Department of Philosophy, History, Culture and Art Studies, University of Helsinki, Finland}

\begin{document}

\maketitle

\begin{abstract}
We study abductive, causal, and non-causal conditionals in indicative and counterfactual formulations  using  probabilistic truth table tasks  under incomplete probabilistic knowledge ($N=80$). We frame the task as a probability-logical inference problem. 
The most frequently observed response
type across all conditions was a class of conditional event interpretations
of conditionals; it was followed by conjunction interpretations. An interesting minority of participants neglected some of the relevant  imprecision involved in the premises when inferring lower or upper probability bounds on the target conditional/counterfactual (``halfway responses''). We discuss the results in the light of \emph{coherence-based probability logic} and the new paradigm psychology of reasoning.

\end{abstract}

\section{Introduction}

The \emph{probabilistic truth table task} was introduced by two independent studies at the beginning of this millennium \cite{evans03,oberauer03}. It can be seen as   one of the starting points of the new (probability-based) paradigm psychology of reasoning \cite<see, e.g.>{baratgin14,elqayam16,oaksford07,over09,pfeifer13b,pfeifer14}, which started to replace the the old (classical logic-based) paradigm psychology of reasoning. The probabilistic truth table task was constructed to investigate how people interpret conditionals (i.e., indicative sentences of the form \emph{If $A$, then $C$}). As its name suggests, the task consists in inferring the degree of belief in a conditional based on probabilistic information attached to the truth table cases. What are truth table cases? Let $A$ and $C$ denote two propositions (i.e., sentences for which it makes sense to assign the truth values \emph{true} or \emph{false}) like \emph{a fair die is rolled} and \emph{the side of the die  shows an even number}, respectively.  The  four truth table cases  induced by  $A$ and $C$ are: $A \wedge C$,  $A \wedge \neg C$,  $\neg A \wedge C$, and $\neg  A \wedge \neg C$,
where $\wedge$ denotes conjunction (``and'') and $\neg$ denotes negation. Classical logic is bivalent, involving only the truth values \emph{true}  and \emph{false}. Therefore, the conditional   defined in classical logic (i.e., the \emph{material conditional}, see Table~\ref{TAB:ttables}) is either true or false. The \emph{conditional event} $C|A$ involved in conditional probability ($p(C|A)$), however,  is not bivalent, as it is \emph{void} (or undetermined) if its antecedent $A$ is false (see Table~\ref{TAB:ttables}). Therefore, it cannot be represented by the means of classical logic.

Table~\ref{TAB:ttables} presents the truth conditions of the most important psychological interpretations for adult reasoning about indicative conditionals (i.e., conditional event, conjunction, and material conditional). Moreover, it presents the biconditional and biconditional event interpretations, which were reported  in developmental psychological studies   \cite<see, e.g.>{barrouillet15}. 

Psychological evidence for the  conditional event interpretation was already observed within the old paradigm psychology of reasoning  \cite<see, e.g.,>{wason72}. The response pattern, which is consistent with the conditional event interpretation  was seen as irrational (dubbed  ``defective truth table''), as it violates the semantics of the material conditional. The material conditional   used to be the  gold standard of reference for the meaning of indicative  conditionals in the old paradigm. However, within the new paradigm psychology of reasoning this response pattern is, of course, perfectly rational  \cite{over17,pfeifertulkki17}.
\begin{table}[!ht]
\begin{center} 
\caption{Truth tables for the material conditional $A \supset C$,
 the conjunction $A\wedge C$,  the biconditional $A\equiv C$, the biconditional event $C||A$ (i.e., $A\wedge C|A\vee C$), and the conditional event $C|A$. } 
\label{TAB:ttables} 
\vskip 0.12in
\begin{tabular}{ccccccc}\hline
$A$ & $C$ & $A \supset C$& $A\wedge C$& $A\equiv C$&$C||A$&$C|A$\\\hline
true &true &true &true &true & true&true  \\
true &false &false &false &false & false&false  \\
false &true &true &false &false&  false &void  \\
false &false &true &false &true&void & void  \\\hline
\end{tabular}
\end{center}
\end{table}

 While classical truth table tasks require to respond truth values, \emph{probabilistic} truth table tasks require to respond degrees of belief. Following \citeA{pfeifer13b}, we interpret  the task   as a probability logical inference problem. Specifically, it is formalised as a probability logical argument with   assigned degrees of belief in the four truth table cases  as its \emph{premises} and the degree of belief in a conditional as its \emph{conclusion}. The inference problem consists in propagating the uncertainties of the premises to the conclusion. As an example, consider the conditional probability  interpretation of the ($p(C|A)$) conditional \emph{If $A$, then $C$} as the conclusion. This argument scheme is formalised by:
\begin{quote}
\begin{itemize}
\item[($\mathcal{A}$)]From $p(A \wedge C)=x_1$,  $p(A \wedge \neg C)=x_2$,  $p(\neg A \wedge C)=x_3$, and $p(\neg  A \wedge \neg C)=x_4$, infer $p(C|A)=x_1/(x_1+x_2)$. 
\end{itemize}
\end{quote}
In argument scheme~($\mathcal{A}$), the fraction $x_1/(x_1+x_2)$ is the probability propagation rule for the conditional probability. For different interpretations of the conditional, the probability propagation rules differ. Table~\ref{TAB:ProbProp} presents the corresponding probability propagation rules of the interpretations given in Table~\ref{TAB:ttables}.

\begin{table}[!ht]
\begin{center} 
\caption{Probability propagation rules for the different interpretations of \emph{If $A$, then $C$}. The premise set $\{p(A \wedge C)=x_1, p(A \wedge \neg C)=x_2,  p(\neg A \wedge C)=x_3, p(\neg  A \wedge \neg C)=x_4\}$ entails the respective conclusion (see also Table~\ref{TAB:ttables}).} 
\label{TAB:ProbProp} 
\vskip 0.12in
\begin{tabular}{ll}\hline
Interpretation &Conclusion \\\hline
Material conditional&$p(A \supset C)=x_1+x_3+x_4$\\
Conjunction& $p(A\wedge C)=x_1$\\
Biconditional & $p(A\equiv C)=x_1+x_4$\\
Biconditional event& $p(C||A)=x_1/(x_1+x_2+x_3)$\\
Conditional event& $p(C|A)=x_1/(x_1+x_2)$ \\\hline
\end{tabular}
\end{center}
\end{table}

 The main empirical result of classical probabilistic truth table tasks is that the dominant responses are consistent with the conditional event interpretation of conditionals. Moreover, if people solve the task several times, some people shift to the conditional event interpretation during the course of the experiment  \cite<see, e.g.,>{fugard11a,pfeifer15a}.  

A key feature of classical probabilistic truth table tasks is that they 
 present \emph{complete probabilistic knowledge} w.r.t.\ the truth table cases: all (point-valued) probabilities $x_1, \dots x_4$ are available to the participant \cite<see, e.g.,>{evans03,oberauer03,fugard11a,pfeifer15a}. When all probabilities involved in   argument scheme~($\mathcal{A}$) are given as point values,  for example, it is possible to   infer a \emph{precise} (point-valued) probability of $C|A$. Of course, $x_3$ and $x_4$ are irrelevant for calculating $p(C|A)$ in this case. However, as full probabilistic information is usually not available in everyday life, we argue for \emph{investigating incomplete probabilistic knowledge}. If $x_1$ in argument scheme~($\mathcal{A}$) is only  available as an \emph{imprecise} (i.e., interval-valued) probability, i.e., $x_1'  \leq p(A\wedge C) \leq x_1''$, the probability of the conclusion of~($\mathcal{A}$) is also imprecise, i.e., 
$x_1'/(x_1'+x_2) \leq p(C|A) \leq x_1''/(x_1''+x_2).$
Table~\ref{TAB:interpretations} (see below) presents  a numerical illustration of the different interpretations of conditionals in the probabilistic truth table task with imprecise premises. Incomplete probabilistic knowledge has  not  been investigated within the probabilistic truth table task paradigm yet \cite<for an exception, where only indicative conditionals were investigated, see>{pfeifer13b}.

With only few exceptions \cite<i.e.,>{over07b,pfeifer15a}, the important classes of causal conditionals  and counterfactuals  have not been investigated yet  within the probabilistic truth table task paradigm. Causal conditionals are characterized by connecting cause (i.e., the conditional's antecedent)  and effect (i.e., the conditional's consequent), like \emph{If you take aspirin, your headache will disappear}. Counterfactuals are conditionals in subjunctive mood, where the grammatical structure signals that the antecedent is false. For instance, \emph{If you were to take aspirin, your headache would disappear}, which signals that you had not taken aspirin yet. This example is a counterfactual version of the above described causal conditional. Of course, there are also non-causal versions of counterfactuals, like \emph{If the side of a  card were to show an ace, it would show spades}.  

Formally, the meaning of counterfactuals can be
modeled by the \emph{coherence-based probability logic} of nested conditionals
\cite{GiSa13c,gilio14,2016:SMPS1,GOPSsubm}. In a nutshell, a counterfactual is interpreted as a ``conditional (if $B$, then $C$) conditionalized on the factual statement ($A$)'', where the factual statement logically contradicts the antecedent of the conditional (i.e., $A \wedge B$ is logically false). Specifically, let
$A$, $B$, and $C$ denote three events, where $A$ is the factual statement which logically contradicts  the antecedent $B$ of the conditional event $C|B$. It has been shown that  the \emph{prevision of the conditional random quantity}
$(C|B)|A$  equals  the \emph{conditional probability of the conditional event} $C|B$ \cite[see Example 1,
p. 225]{GiSa13c}. Thus, the counterfactual \emph{If $A$ were the case,
  $C$ would be the case} can be modeled by the degree of belief in the
conditional random quantity $(C|A)|\neg A$ which equals to $p(C|A)$
(i.e., $Prevision((C|A)|\neg A)=p(C|A)$). Therefore, we hypothesize that the participants' degrees of belief in counterfactuals are equal to  corresponding conditional
probabilities.

To our knowledge, \emph{abductive conditionals} have not yet been investigated in the probabilistic truth table task paradigm. Abductive conditionals can be conceived as reversed causal conditionals, characterized as follows: the effect is located in the conditional's antecedent and the   cause is located in the conditional's  consequent. For example, \emph{If your headache  disappeared, then you took aspirin}. Abductive inferences are also known as inferences to the best explanation \cite<for philosophical and psychological overviews on abduction see, e.g.,>[respectively]{sep-abduction, lombrozo12}.  Like indicative and causal conditionals, abductive conditionals can be formulated in indicative and in subjunctive mood. 

The aim of the present study is to help to fill the above mentioned research gaps. Specifically, we aim to shed light on the following questions using  probabilistic truth table tasks under incomplete probabilistic knowledge:
\begin{itemize}
\item Are there reasoning strategies for inferring lower and upper bounds in the context of incomplete probabilistic knowledge?
\item Is the conditional event interpretation dominant for abductive, causal,  and non-causal counterfactuals?

\end{itemize}




\section{Method}

\paragraph{Materials and Design}
The task material consisted of 18 pen and paper tasks, preceded by four examples explaining the answer format. The task sequence consisted of nine different tasks that were presented twice in the same random order (i.e., task T10 is a  repetition of task T1). The tasks were designed to test how participants infer about uncertain conditional sentences in four different situations (see  Table~\ref{Design}). For the first two conditions we used a non-causal scenario in both indicative and counterfactual moods. For the other two conditions we  used two variations of a causal scenario in counterfactual mood; inference from causes to effects (causal) and inference from effects to causes (abductive). The material was adapted from  probabilistic truth table tasks, which  involved precise premises \cite{fugard11a,pfeifer15a}. 

\begin{table}[!ht]
\begin{center} 
\caption{Experimental conditions C1--C4 defined by the types and formulations of conditionals, and sample sizes.} 
\label{Design} 
\vskip 0.12in
\begin{tabular}{cllc} 
\hline
  & Type                 &  Formulation & Sample \\
\hline
C1 & non-causal & indicative & ($n_1=20$) \\
C2 & non-causal & counterfactual & ($n_2=20$) \\
C3 & causal & counterfactual & ($n_3=20$) \\
C4 & abductive & counterfactual & ($n_4=20$) \\
\hline

\end{tabular} 
\end{center} 
\end{table}

\begin{figure}[ht]
\begin{center}$
\ifx\JPicScale\undefined\def\JPicScale{0.2}\fi
\psset{unit=\JPicScale mm}
\psset{linewidth=0.3,dotsep=1,hatchwidth=0.3,hatchsep=1.5,shadowsize=1,dimen=middle}
\psset{dotsize=0.7 2.5,dotscale=1 1,fillcolor=black}
\psset{arrowsize=1 2,arrowlength=1,arrowinset=0.25,tbarsize=0.7 5,bracketlength=0.15,rbracketlength=0.15}
\begin{array}{cccccc}
\begin{pspicture}(0,0)(40,40)
\pspolygon[](0,0)(40,0)(40,40)(0,40)
\rput{0}(20,20){\psellipse[fillstyle=solid](0,0)(10,-10)}
\end{pspicture}
&
\begin{pspicture}(0,0)(40,40)
\pspolygon[](0,0)(40,0)(40,40)(0,40)
\pspolygon[fillstyle=solid](10,30)(30,30)(30,10)(10,10)
\end{pspicture}
&
\begin{pspicture}(0,0)(40,40)
\pspolygon[](0,0)(40,0)(40,40)(0,40)
\pspolygon[fillstyle=solid](10,30)(30,30)(30,10)(10,10)
\end{pspicture}
&
\begin{pspicture}(0,0)(40,40)
\pspolygon[](0,0)(40,0)(40,40)(0,40)
\pspolygon[](10,10)(30,10)(30,30)(10,30)
\end{pspicture}
&
\begin{pspicture}(0,0)(40,40)
\pspolygon[](0,0)(40,0)(40,40)(0,40)
\pspolygon[](10,10)(30,10)(30,30)(10,30)
\end{pspicture}
&
\begin{pspicture}(0,0)(40,40)
\pspolygon[](0,0)(40,0)(40,40)(0,40)
\put(10,10){\makebox(20,20)[cc]{?}}
\end{pspicture}

\end{array}$
\end{center}\vspace{-.4cm}
\caption{Example of  presented sides of a die (non-causal task). Covered sides  are indicated by the question mark.} 
\label{FIG:SampleDice}
\end{figure}
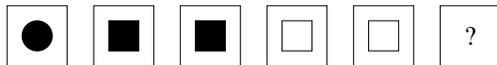

For the \emph{non-causal scenario}  we used a vignette story about a six-sided die. The story describes that the die was randomly thrown so that the participants did not know which of the sides ended up facing upwards. The sides of the die were illustrated as six squares. Each side had an image of a black or white geometric figure. In tasks T1, T2, T10, and T11 all sides of the die were shown. To introduce  incomplete probabilistic knowledge we presented ``covered'' sides in the rest of the tasks. Covered sides  were indicated by a question mark. Figure~\ref{FIG:SampleDice} shows an example of how we presented the six sides of a die.

\begin{figure}[ht]
\begin{center}
\includegraphics[width=.3\textwidth]{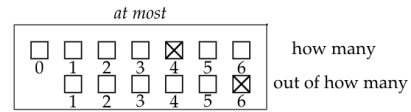}
\end{center}\vspace{-.4cm}
\caption{Upper bound response  (``at most 4 out of 6'').} 
\label{FIG:Tickbox}
\end{figure}

Next, the participants were presented with the question ``How sure can you be that the following sentence holds?'' (\emph{Kuinka varma voit olla siit\"a, ett\"a seuraava lause pit\"a\"a paikkaansa?}). The target sentences were highlighted with a frame to make the scope of the question clear,  for example:
\begin{center}
\noindent
\fbox{
  \parbox{.43\textwidth}{\footnotesize {\bf If} the figure on the 
upward facing side of the die is a \emph{circle},\newline {\bf then} the figure is \emph{black}.
  }
}
\end{center}
The answer format had two sets of tick boxes in a ``x out of y'' format for responding interval-valued degrees of belief. The two response boxes  were labeled accordingly (``at least'' and ``at most''). It was explained in the introduction   to give point valued responses by marking the same numbers in both response boxes (i.e., lower and upper bounds coincide).
See Figure~\ref{FIG:Tickbox} for an example.

The target sentence in the non-causal tasks was formulated in terms of  ``If $A$, then $C$''. In all non-causal tasks the antecedent mentioned a form and the consequent mentioned a color. After completing each task, the participants were asked to rate their   confidence in the correctness of their response on a 10-step rating scale from ``fully confident that your answer is incorrect" to ``fully confident that your answer is correct".

Apart from the following two differences, the counterfactual task version was identical to the indicative version of the task: (i) we added a  factual statement which contradicted the antecedent of the target sentence and (ii) the target sentence was formulated in subjunctive mood. ``The form of an upward-facing side of the die is a cube'' is an example of a factual statement and the corresponding  target sentence is:  ``if the figure on the upward facing side of the die were a circle, then the figure would be black'' (\emph{Jos yl\"osp\"ain osoittavan kyljen kuvio ol\underline{isi} ympyr\"a, niin se kuvio olisi musta}). The   suffix \emph{-isi} in the  Finnish original indicates the counterfactual mood.

\begin{figure}[ht]
\begin{center}
\includegraphics[width=.45\textwidth] {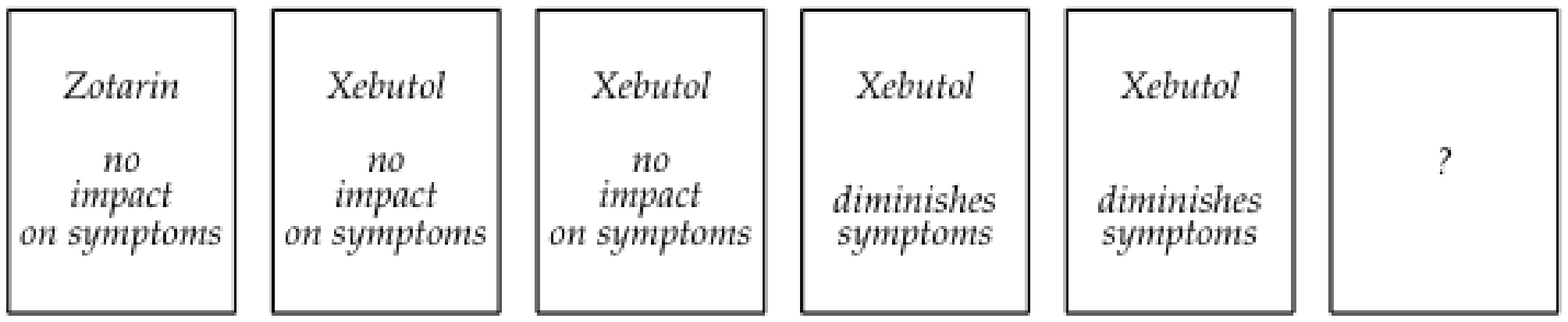}
\end{center}\vspace{-.5cm}
\caption{Example of presented patient reports (causal / abductive tasks; see also Figure~\ref{FIG:SampleDice}).} 
\label{FIG:ExampleReports}
\end{figure}

For the \emph{causal} and \emph{abductive} conditions, the tasks were structurally identical to the (non-causal) dice-scenario.  However, instead of dice, a vignette story about drugs and their effects  created a  causal  scenario. In the vignette
story, six patient reports were shown to the participants.  The patient reports were illustrated as six rectangles having a name of a fictional drug and a result of the medication (either ``diminishes symptoms'' or ``no impact on the symptoms'').
We used question marks on some patient reports (like in the dice scenario) to introduce incomplete probabilistic knowledge.  Figure~\ref{FIG:ExampleReports} shows an example of the patient reports. This example contains the same probabilistic information as the die-example  in Figure~\ref{FIG:SampleDice}.


In the causal version of the task material the antecedent denotes the name of a drug and the consequent denotes an effect. In the abductive version the order was reversed: first an effect was presented and then a drug was named. In this way the tasks called for either \emph{causal inferences} from  causes to effects, or \emph{abductive inferences} from  effects to  causes. As the material was formulated in counterfactual mood, we added  a factual statement to each task, which  contradicted the antecedent of the target sentence. 

\paragraph{Participants and Procedure}
Eighty students from the University of Helsinki (Finland) participated in the experiment. The students were native Finnish speakers with no previous academic training in logic or probability. Each participant was tested individually. The paper and pencil tasks were followed by a short structured interview about how the participants had interpreted the target tasks. Participants were paid 15{\EUR}  for their participation.

\section{Results and Discussion}
We performed Fisher's exact tests to investigate whether the four different versions of the task booklets had an impact on the participants' degrees of belief in the respective target sentences. After p-value corrections for multiple significance tests,  we did not observe  significant differences between the four conditions and we therefore  pooled the data.

Table~\ref{Results1} shows the percentages of responses according to the different interpretations of conditionals. All tasks differentiate between the conditional probability, the conjunction, and  the material conditional interpretation. A subset of the tasks differentiates between biconditional and biconditional event interpretations as well.
 Conditional probability interpretations  marked with indices, however,  were patterns identified from the data and were not anticipated during the construction of the task material. Therefore,  not all tasks differentiate among all interpretations. Table~\ref{TAB:interpretations} shows the normative answers for each interpretation for the example task discussed  in the previous section (see Figure~\ref{FIG:SampleDice}). Both, lower and upper  bound responses, had to match the normative lower and upper bounds  for the categorization of the responses in Table~\ref{Results1}. Since each response box enables 42 different ``$X$ out of $Y$'' responses,  and since both, lower and upper  bound responses needed to match for the classification, the \emph{a priori} chance for guessing an interpretation was very low (i.e.,   $1/(42^2) = 0.0006$).

\begin{table}[!ht]
\begin{center} 
\caption{Example of predicted responses where the task consists in inferring the degree of belief in the conditional ``{\bf If} the figure on the 
upward facing side of the die is a \emph{circle}, {\bf then} the figure is \emph{black}'' (i.e., the conclusion), based on the  die presented in Figure~\ref{FIG:SampleDice} (i.e., Figure~\ref{FIG:SampleDice} contains the premises). The index $\overline{l}$ (resp., $\overline{u}$) denotes conditional probability responses where the covered sides are ignored for inferring the lower (resp., upper) bound response. These response types are the ``halfway lower'' and ``halfway upper'' interpretations, respectively. $\overline{lu}$ denotes conditional probability responses where covered sides are ignored for inferring both bound responses, i.e., the ``halfway both interpretation''.  See also Table~\ref{TAB:ProbProp}.} 
\label{TAB:interpretations}
\vskip 0.12in
\begin{tabular}{lcc} 
\hline
                      Interpretation          & \multicolumn{2}{c}{Predictions}\\ 
                  &  at least & at most \\
\hline
$p(\text{black }|\text{ circle})$                  &1 out of 2  & 2 out of 2 \\
$p(\text{black }|\text{ circle})_{\overline{l}}$                    & 1 out of 1  & 2 out of 2\\
$p(\text{black }|\text{ circle})_{\overline{u}}$                    & 1 out of 2 & 1 out of 1 \\
$p(\text{black }|\text{ circle})_{\overline{lu}}$                   & 1 out of 1 & 1 out of 1\\
$p(\text{circle }\wedge\text{ black})$                    & 1 out of 6   &  2 out of 6  \\
$p(\text{circle }\supset\text{ black})$                  & 5 out of 6   & 6 out of 6     \\
$p(\text{circle }\equiv\text{ black})$                   & 3 out of 6   & 4 out of 6     \\
$p(\text{circle }||\text{ black})$                 & 1 out of 4   &  2 out of 4     \\
\hline
\end{tabular} 
\end{center} 
\end{table}

\begin{table}[!ht]
\begin{center} 
\caption{Percentages of responses from all four conditions and all 18 tasks (T1--T18; $N=80$).  ``Grouped $p(\cdot|\cdot)$'' denotes the sum  of all conditional probability responses, including those marked with the indices. The halfway interpretations (indices $\overline{l}$ and $\overline{u}$) and the   numerical predictions are explained in Table~\ref{TAB:interpretations}.  ``- -'' denotes cases where  different conditional probability interpretations cannot be individuated (i.e., in the point value tasks). Similarly, {``[-~-]''} denotes cases where  biconditional and biconditional event  interpretations cannot be distinguished from the other interpretations.  The interpretations are explained in Table~\ref{TAB:ProbProp}.  ``$^{\star}$'' denotes psychological main interpretations.} 
\label{Results1} 
\vskip 0.12in
\begin{tabular}{lcccccc} 
\hline
Interpretation                  &  T1 & T2 & T3 & T4 & T5 & T6 \\
\hline
$[p(\cdot|\cdot)]^{\star}$                  & [46]   &  [55]  & [15]  & [19]   & [24]   & [24]   \\
$[p(\cdot|\cdot)_{\overline{l}}]$                    & [- -]  & [- -]   & [5]  & [13]  &[18]    & [11]   \\
$[p(\cdot|\cdot)_{\overline{u}}]$                    & [- -]  & [- -]   & [23] & [10]  & [13]   & [11]   \\
$[p(\cdot|\cdot)_{\overline{lu}}]$                   &  [- -] & [- -]    & [0] &  [3] & [1]   &[0]    \\
Grouped $p(\cdot|\cdot)$&  46   & 55   & 43  & 44   & 55   &    46\\[.2em]
$p(\cdot\wedge\cdot)^{\star}$                    & 29    &  28  & 34   & 39   & 34   & 31   \\
$p(\cdot\supset\cdot)^{\star}$                  & 1    & 0   & 0   &  0  &  0  & 1   \\
$p(\cdot \equiv \cdot)$                   & [- -]    & [- -]   & 1   &  [- -]  & [- -]  & 0   \\
$p(\cdot ||\cdot)$                  & [- -]    & [- -]   & 3   &  [- -] &  [- -]  & 0   \\
Other                   &  24   &  18  &  20  & 18   &  11  & 21   \\
\hline
                     &  T7 & T8 & T9 & T10 & T11 & T12 \\ 
\hline
$[p(\cdot|\cdot)]^{\star}$                  & [24]   &  [28]  & [26] & [44]  &  [55]  & [25]   \\
$[p(\cdot|\cdot)_{\overline{l}}]$                    & [10]  & [15]   & [10] &  [- -] & [- -]   & [9]   \\
$[p(\cdot|\cdot)_{\overline{u}}]$                    & [16]  & [8]    & [10] & [- -]  & [- -]   &  [23]  \\
$[p(\cdot|\cdot)_{\overline{lu}}]$                   &  [0] & [0]    & [0] &  [- -] & [- -]   &[0]    \\
Grouped $p(\cdot|\cdot)$&  46   & 55   & 43  & 44   & 55   &    46\\[.2em]
$p(\cdot\wedge\cdot)^{\star}$                    &  34   &  29  & 33   &  26   & 29    & 30   \\
$p(\cdot\supset\cdot)^{\star}$   & 0    & 0   & 0   &  1  &  0  & 0   \\
$p(\cdot \equiv \cdot)$                   & [- -]    & [- -]   & [- -]   &  [- -]  &  [- -]  & 0   \\
$p(\cdot ||\cdot)$            &  [- -]   &  [- -]  & [- -]  &  [- -]   &  [- -]   &  0  \\
Other                    & 16    & 21   &  21  &  18   & 18    & 14   \\
\hline
                     &  T13 & T14 & T15 & T16 & T17 & T18 \\
\hline
$[p(\cdot|\cdot)]^{\star}$                  & [34]   & [33]   & [29]  & [26]  & [31]   & [31]    \\
$[p(\cdot|\cdot)_{\overline{l}}]$                    & [9]  & [13]   & [11] & [10]  &  [18]  &  [13]  \\
$[p(\cdot|\cdot)_{\overline{u}}]$                    & [11]  & [10]   &[11]  & [15]  & [8]   & [11]   \\
$[p(\cdot|\cdot)_{\overline{lu}}]$                   &  [0] & [0]    & [1] &  [3] & [0]   &[0]    \\
Grouped $p(\cdot|\cdot)$&  46   & 55   & 43  & 44   & 55   &    46\\[.2em]
$p(\cdot\wedge\cdot)^{\star}$                    &  28    & 30    &  26   & 29    & 25    & 28   \\
$p(\cdot\supset\cdot)^{\star}$                  & 0     &  0   &  0   &  0   &  0   &  0  \\
$p(\cdot \equiv \cdot)$                  & [- -]    & [- -]   & 0   &  [- -]  &  [- -]  & [- -]   \\
$p(\cdot ||\cdot)$                  & [- -]    & [- -]   & 3  &  [- -]  &  [- -]  & [- -]   \\
Other                    & 19     &  15   &  19   &  18   & 19    &  18  \\
\hline

\end{tabular} 
\end{center} 
\end{table}

\begin{figure}[!ht]
\begin{center}
\includegraphics[width=.4\textwidth] {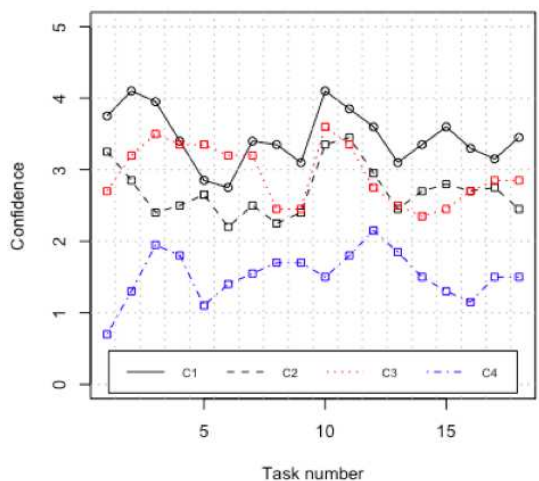}
\end{center}\vspace{-.45cm}
\caption{Mean confidence values for tasks T1--T18 by condition. C1--C4 denote the four condition as defined in Table~\ref{Design}.}
\label{FIG:Confidence}
\end{figure}

The task material was designed such that  the normative predictions  of the three main psychological interpretations of conditionals (i.e.,  conditional probability, conjunction, and material conditional) were unique for each task. During the analysis we identified three further response strategies related to the conditional probability interpretation. In what we call \emph{halfway lower} interpretation (denoted by $p(\cdot |\cdot)_{\overline{l}}$) the upper bound  is the same as in conditional probability, but the lower bound response differs in that the covered sides (i.e., sides marked with question mark) are ignored. \emph{Halfway upper} interpretation (denoted by $p(\cdot |\cdot)_{\overline{u}}$) is the same, but in reverse order. In a \emph{halfway both} interpretation the covered sides are ignored for both bound responses. As these answer strategies are in a sense partial versions of the conditional probability interpretation, we combined their results with conditional probability answers into \emph{grouped conditional probability}. Notice that the tasks T1,T2, T10 and T11 with full information (i.e., no question marks) have the same value for lower and upper bound answers. Therefore the halfway responses could not be distinguished from the conditional probability answers in these tasks.

Of all 1440 responses, 32.1\% were consistent with conditional event responses, 29.9\% were consistent with conjunction responses, and 0.2\% were consistent with material conditional responses. Like the material conditional, also the biconditional and the biconditional event response frequencies play a neglectable r\^ole in the data (0\%--3\% in the four tasks T3, T6, T12, T15 where these interpretations were differentiated). The predominant response strategy in point-valued tasks (T1, T2, T10 and T11) was consistent with the conditional probability interpretation. In nine out of 14 tasks with incomplete probabilistic information (i.e., tasks involving ``covered'' sides or patient reports) the predominant answer strategy was conjunction. We also observed shifts of interpretation towards conditional probability: comparing the first three tasks with incomplete information (i.e., T3--T5) to the last three (i.e., T16--T18), the number of conditional probability answers increased from 19\% to 30\%, and conjunction answers decreased from 35\% to 27\%. This replicates  shifts of interpretations reported in the literature \cite{fugard11a,pfeifer13b}. 


However, when all the conditional probability response types are grouped together, the resulting set of response strategies is clearly the predominant one in all tasks. 51.5\% of all answers are consistent with the grouped conditional probability responses.  18.1\% were ``other'' responses, that is, responses that did not fit the grouped conditional probability, conjunction, biconditional, biconditional event, or material conditional. Thus, in total 81.9\% of the data can be modeled by the investigated hypotheses concerning the interpretation of conditionals.

Compared to a previous study that investigated non-causal indicative conditionals under incomplete probabilistic information  \cite[i.e., similar tasks as in condition C1]{pfeifer13b}, our results show lower level of conditional event responses (compared to the previous 65.6\%), and higher levels of conjunction responses (compared to the previous 5.6\%). The material conditional responses were similar (previously 0.3\%). \citeA{pfeifer15a}  investigated conditionals under complete probabilistic knowledge and  used similar tasks as in our conditions C1, C2 and C3. These authors also reported higher levels of conditional probability answers and lower levels of conjunction responses, while material conditional responses were similarly low. The lower levels of conditional probability responses may be explained by the apparent higher complexity of the tasks used in the present experiment. The tasks are more complex (i) because of the combination of using  counterfactuals as target sentences in three of four conditions and (ii) because of imprecise probabilities in the premises (i.e., incomplete probabilistic information). 

The tendency to give answers that partially coincide with conditional probability has also been found in a previous study which tested non-causal cases in indicative mood with similar task material as we used for condition C1 \cite{pfeifer13b}. In that study our halfway lower-interpretation is referred to as ``halfway conditional event strategy''. However, in the present study we found two completely new strategies: the halfway upper-  and the halfway both-interpretation. The halfway upper-interpretation is particularly interesting, as it explains 12.8\% of the total 1120 responses,  slightly more than the halfway lower response strategy (i.e., 11.6\%). Halfway conditional probability responses might  unburden the working memory load by  ignoring the covered sides \cite<see also>{pfeifer13b}.

 Figure~\ref{FIG:Confidence} shows the results of the confidence ratings. We performed analyses of variance to investigate impacts of the different conditions. After Holm-Bonferroni corrections we observed significant differences within the three tasks  T1, T2 and T10. The corresponding p-values were 0.006, 0.01, and 0.008. In each of these tasks---as well as in all other tasks---the condition C4 had lower mean confidence values compared to the other conditions. The lower confidence may be because of the apparent higher difficulty of  the task material in condition C4 for  two reasons: first, the target sentence was a  counterfactual. Because of the inconsistency between the factual statement and the antecedent, many participants reported counterfactuals as puzzling in the post-test interview. Second, abductive tasks required ``backward'' inference from effects to causes and are incongruent with the more natural \emph{if cause, then effect}-direction. In general, backward inferences  are known  to be harder to draw compared to forward inferences \cite{evans81}. 

\section{Concluding Remarks}
We investigated how people reason about conditionals under incomplete probabilistic knowledge. The novel features in our test design were comparisons of \emph{causal} and \emph{abductive} scenarios, as well as \emph{counterfactuals} under \emph{incomplete} probabilistic knowledge.
One of our main findings is that the dominant response is consistent
with the conditional event interpretation of conditionals
among all four groups. Moreover, we discovered  two major answer strategies, \emph{halfway upper} and \emph{halfway lower conditional event responses}, which can be understood as strategies to unburden the working memory load.

Inferentialist accounts of conditionals propose that there should be an inferential relation between the antecedent and the consequent \cite<see, e.g.,>{douven2016}. Thus, when conditionals with inferential  relations (e.g., causal  or abductive ones) are compared with conditionals where no apparent inferential relation exists (like in our conditions C1 and C2), one would expect significant differences. Our data, however, do not support this inferentialist hypothesis. 

The results of our paper broaden the area of inferences where  conditional probability seems to be the best predictor for how people reason.
We have shown that coherence-based probability logic provides a formalization of the meaning of counterfactuals and provides a rationality framework for reasoning under complete and incomplete probabilistic knowledge. This suggests that it may also be suitable for subclasses of causal reasoning like abductive reasoning, which is important in the studies on (scientific) explanation and learning.

{\small\paragraph{Acknowledgments}
This research was supported by the DFG project
    PF~740/2-2 (awarded to Niki Pfeifer) as part of the Priority Program ``New Frameworks of Rationality'' (SPP1516).}




\end{document}